\documentclass[runningheads]{llncs}
\usepackage{graphicx}
\usepackage{comment}
\usepackage{amsmath,amssymb} 
\usepackage{color}

\usepackage[width=122mm,left=12mm,paperwidth=146mm,height=193mm,top=12mm,paperheight=217mm]{geometry}

\usepackage{subfig, mathtools, caption, float}
\usepackage[pagebackref=true,breaklinks=true,letterpaper=true,colorlinks,bookmarks=false]{hyperref}

\usepackage{color}
\usepackage{colortbl}
\newcommand{\bgGray}[1]{\cellcolor[gray]{0.85} #1}

\newcommand{\ie}{\textit{i}.\textit{e}.}
\newcommand{\eg}{\textit{e}.\textit{g}.}

\begin{document}
\pagestyle{headings}
\mainmatter
\def\ECCVSubNumber{XXXX}  

\title{Author Guidelines for ECCV Submission} 
\title{Learning to Predict Context-adaptive Convolution for Semantic Segmentation} 

\titlerunning{\ }
%
\author{Jianbo Liu\inst{1} \and
Junjun He\inst{2} \and
Jimmy S. Ren\inst{3} \and
Yu Qiao\inst{2} \and
Hongsheng Li\inst{1}}
%
\authorrunning{\ }
%
\institute{CUHK-SenseTime Joint Laboratory, The Chinese University of Hong Kong \and
Shenzhen Key Lab of Computer Vision and Pattern Recognition, Shenzhen Institutes of Advanced Technology,
Chinese Academy of Sciences\and
SenseTime Research \\
\email{\{liujianbo@link, hsli@ee\}.cuhk.edu.hk}}
\maketitle


\begin{abstract}
Long-range contextual information is essential for achieving high-performance semantic segmentation. 
Previous feature re-weighting methods \cite{Zhang_2018_CVPR} demonstrate that using global 
context for re-weighting feature channels can effectively improve the accuracy
of semantic segmentation. However, the globally-sharing feature re-weighting
vector might not be optimal for regions of different classes in the input image. 
In this paper, we propose a Context-adaptive Convolution Network (CaC-Net) to predict a 
spatially-varying feature weighting vector for each spatial location of the semantic feature maps. 
In CaC-Net, a set of context-adaptive convolution kernels are predicted from the global contextual information in a parameter-efficient manner. 
When used for convolution with the semantic feature maps, the predicted convolutional kernels can generate the spatially-varying feature weighting factors capturing both global and local contextual information. 
Comprehensive experimental results show that our CaC-Net achieves superior segmentation performance on three public datasets, PASCAL Context, PASCAL VOC 2012 and ADE20K.
\end{abstract}

\section{Introduction}
Semantic segmentation aims at estimating a category label for each pixel of an input image,
which is a fundamental problem in computer vision. 
It plays an important role in many applications including autonomous driving, image editing, computer-aided diagnosis, etc.
Recently, state-of-the-art approaches leverage the Fully Convolutional Network
({FCN}) \cite{long2015fully} as a base network to encode dense semantic
representations from the input image and predict the class label for each pixel \cite{chen2017deeplab,chen2017rethinking,zhao2017pyramid,Zhang_2018_CVPR,he2019adaptive,takikawa2019gated}.
However, the significant scale variations of objects belonging to the same class and the similar appearances of objects belonging to different classes pose great challenges for semantic segmentation methods. 

To overcome these challenges, many works have been proposed to 
exploit the long-range contextual information from the dense semantic
representations \cite{chen2017deeplab,zhao2017pyramid,Zhang_2018_CVPR,he2019adaptive} 
to better distinguish scale and appearance ambiguity. 
One common solution is to aggregate the local context from different locations or sub-regions.
The contextual information is collected by taking a fixed or an adaptively weighted aggregation from the local neighborhoods \cite{zhao2017pyramid,Zhang_2018_CVPR,fu2019dual,Zhang_2019_CVPR,Li_2019_ICCV}. 

\begin{figure}[t]
\begin{center}
\subfloat{\includegraphics[width=0.96\textwidth]{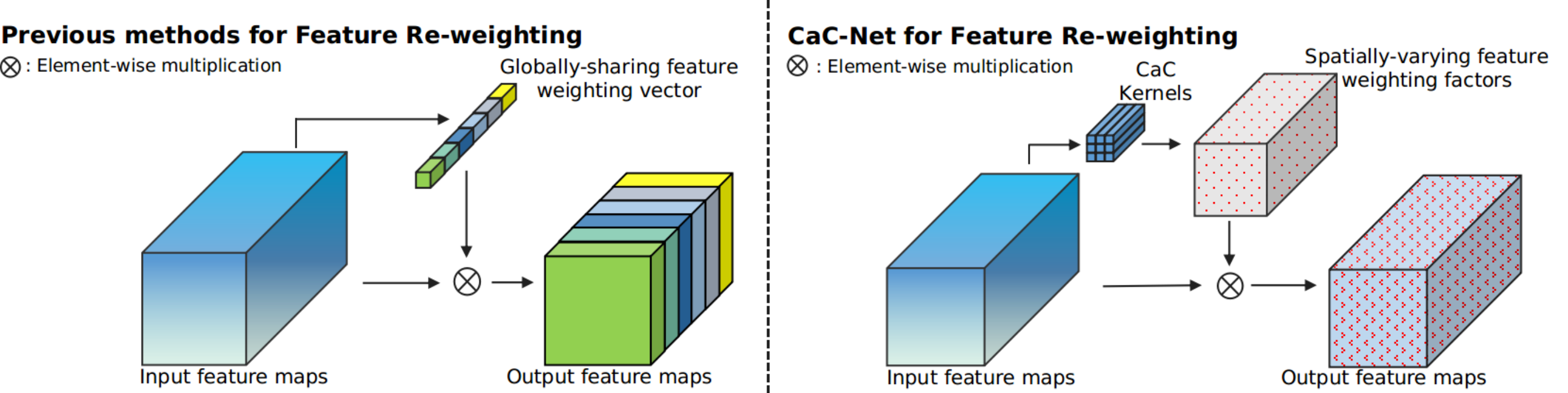}}
\end{center}
\caption{(Left): Previous feature re-weighting methods \cite{Zhang_2018_CVPR,hu2018squeeze} used a globally-sharing
weighting vector to improve the performance of semantic segmentation, but
ignores the spatially varying characteristics of the input image. (Right): Our CaC-Net learns to predict context-adaptive convolution kernels from the
global context for weighting feature channels in a spatially-varying manner}

\label{fig:intro}
\end{figure}

In contrast to encoding the relationship of local spatial context, another category 
of methods attempted to capture the global contextual information for estimating the 
channel-wise feature importances and using it to re-weight the feature channels for improving the 
segmentation accuracy. As illustrated in Figure \ref{fig:intro}, the feature weighting can 
automatically highlight the features that are more relevant to the given scene and suppresses the irrelevant feature. 
For instance, boats usually appear in the sea or rivers but not in indoor
environments. With the global context of a water scene, water-related feature
channels should be higher weighted to increase the probability of predicting boat pixels.
SE-Net \cite{hu2018squeeze} proposed the squeeze-excitation operation to learn a global feature weighting vector for weighting feature maps to improve image classification, detection and segmentation. EncNet \cite{Zhang_2018_CVPR} designed a global context encoding layer to weight features for improving semantic segmentation. 
Although this strategy has shown improved accuracy on segmentation, one of its key problems is 
that all the spatial locations share a common channel-wise weighting vector to calibrate 
the contributions of feature channels. For instance, features in the sky regions and the water regions of a water scene should be weighted differently.

To tackle this challenge, as shown in Figure \ref{fig:intro}, we propose a
novel segmentation framework, which properly weights feature channels with the global context 
but in a spatially varying manner, \ie, feature channels at different spatial
locations are modulated differently based on predictable and input-variant convolutions. 
Although there is a naive solution that simply generates kernels with a large number of parameters to conduct dot-product with the original feature maps, such strategies introduce too many parameters and cannot be used in practice. 
We propose a novel approach, which learns to predict
Context-adaptive Convolution (CaC) kernels from the global context for generating the spatially-varying feature weighting factors.
To reduce the parameters and the computational burden, we do not predict all CAC kernel parameters with fully-connected (FC) layers as previous dynamic kernels \cite{jia2016dynamic} do, which need too many learnable FC parameters. 
We instead propose an efficient context-adaptive kernel learning scheme that
generates the CaC kernel parameters via simple matrix multiplication. The
CaC kernels not only fully encode global context of the input feature maps but also generate context-aware spatially-varying feature weighting factors for each spatial location via depth-wise convolution with the input feature maps. 
In addition, we utilize a series of dilated depth-wise convolutions with different dilation factors to effectively capture information of multiple scales.

Our main contributions can be summarized as threefold:
(1) To better regularize the semantic segmentation with global
        contextual information, we propose to estimate spatially-varying
        feature weighting factors for properly weighting features at different
        spatial locations to improve the semantic segmentation performance.
(2) We train an efficient deep network to predict the context-adaptive
        convolution kernels from the global context of the input image. The
        proposed CaC kernel prediction is  both computational and memory efficient, 
        which can effectively generate spatially-varying weights based on
        both global context and local information.
(3) With the proposed CaC module, our proposed approach achieves state-of-the-art performance on multiple public benchmarks, including PASCAL-Context,
        PASCAL VOC 2012, and ADE20K.

\section{Related Work}
\noindent \textbf{Context aggregation.}\
Fully convolution network \cite{long2015fully} based methods have made great
achievements in semantic segmentation.
With a series of convolution and down-sampling operations, the features of deeper layers gradually capture information with larger receptive fields. However, they still have limited receptive fields and cannot effectively take advantages of the global or long-range context.
Global or long-range contextual information aggregation has been shown their
effectiveness on improving the segmentation accuracy of large homogeneous semantic regions or
objects with large scale variations. 
ParseNet \cite{liu2015parsenet} proposed to capture the global context by concatenating a global pooling feature with the original feature maps.
PSPNet \cite{zhao2017pyramid} designed a Spatial Pyramid Pooling (SPP) module to collect contextual information of different scales.
Atrous Spatial Pyramid Pooling (ASPP) \cite{chen2017deeplab,chen2017rethinking} applied a set of different 
dilated convolutions to capture multi-scale contextual information.
However, these methods treat all pixels in each sub-region with uniform or fixed weights for feature aggregation.

%
To achieve adaptive and flexible feature aggregations, APCNet \cite{he2019adaptive}
proposed an adaptive context module (ACM) to leverage local and global
representations to estimate inter-pixel affinity weights for feature aggregation. 
CFNet \cite{Zhang_2019_CVPR} designed an aggregated co-occurrent feature (ACF) module to
aggregate the co-occurrent context using the pair-wise similarities
in the feature space.
PSANet\cite{zhao2018psanet} aggregated contextual information for each pixel with a predicted
attention map. 
DANet \cite{fu2019dual} proposed to apply a position attention module and channel
attention module with the self-attention mechanism to aggregate features from
spatial and channel dimensions respectively.
%
These techniques show robustness to shape or scale variations of objects and are able to boost the segmentation performance.
However, it is still challenging for this kind of methods to efficiently and accurately find all the pixels belonging to the same object class.

\vspace{5pt}
\noindent
\textbf{Channel-wise feature re-weighting.}\
To take advantages of the global contextual information of the input images, some pioneering methods \cite{hu2018squeeze,Zhang_2018_CVPR} 
have been proposed to re-weight different channels of the 2D feature maps with a scaling vector learned from the global context feature vector. 
Both SE-Net \cite{hu2018squeeze} and EncNet \cite{Zhang_2018_CVPR} learned a globally-sharing attention vector from the global context. 
SE-Net \cite{hu2018squeeze} proposed to learn feature weighting factors by
the squeeze-excitation operation. The squeeze operation aggregates the feature maps across all spatial locations to produce a global context-encoded feature vector. Then the excitation
operation learns the weighting factors from the global-context features.
EncNet \cite{Zhang_2018_CVPR} predicted one globally-sharing feature re-weighting vector using a context encoding module. This module integrates dictionary learning and residual encoding components to learn a global context encoded feature vector, based on which, the feature weighting factor vector is predicted.
However, both methods only consider the global context and output a feature re-weighting vector that is shared across all spatial locations.
Such a globally-sharing weighting scheme might not be suitable for different spatial regions belonging to different objects in the same scene.
To tackle this challenge, we propose to predict learnable kernels for re-weighting the feature maps in a spatially-varying manner. 

\vspace{5pt}
\noindent
\textbf{Dynamic filters.}\
Dynamic filters or kernels were proposed by \cite{jia2016dynamic} to generate content-aware filters,
which are conditional on the input images. These filters are
adaptive to the input and are predicted by the neural network.
It has shown effectiveness in various computer vision applications.
Some methods adopted the predicted dynamic filters for low-level vision and video understanding, \eg, video
interpolation \cite{niklaus2017video} and image denoising \cite{mildenhall2018burst}.
Pixel-Adaptive Convolution (PAC) Network \cite{su2019pixel} was proposed to predict
spatially-varying kernels for several computer vision applications including deep joint image 
upsampling, semantic segmentation and efficient CRF inference.
%
SAC \cite{zhang2017scale} utilized a scale regression layer to predict position-adaptive scale coefficients, which were used to automatically adjust the sizes of receptive fields for objects of different sizes. 
DMNet \cite{he2019dynamic} exploited a set of dynamic filters of different
sizes, which were generated from multi-scale
neighborhoods for handling the scale variations of objects for semantic segmentation.
%
The previous methods with dynamic filters directly
predict all the filter parameters, which are time inefficient and occupy too much memory.
In comparison, our work learns to predict the context-adaptive convolution kernels from global 
context in a parameter-efficient manner.
%

\begin{figure*}[!t]
        \begin{center}
                \includegraphics[width=0.98\textwidth]{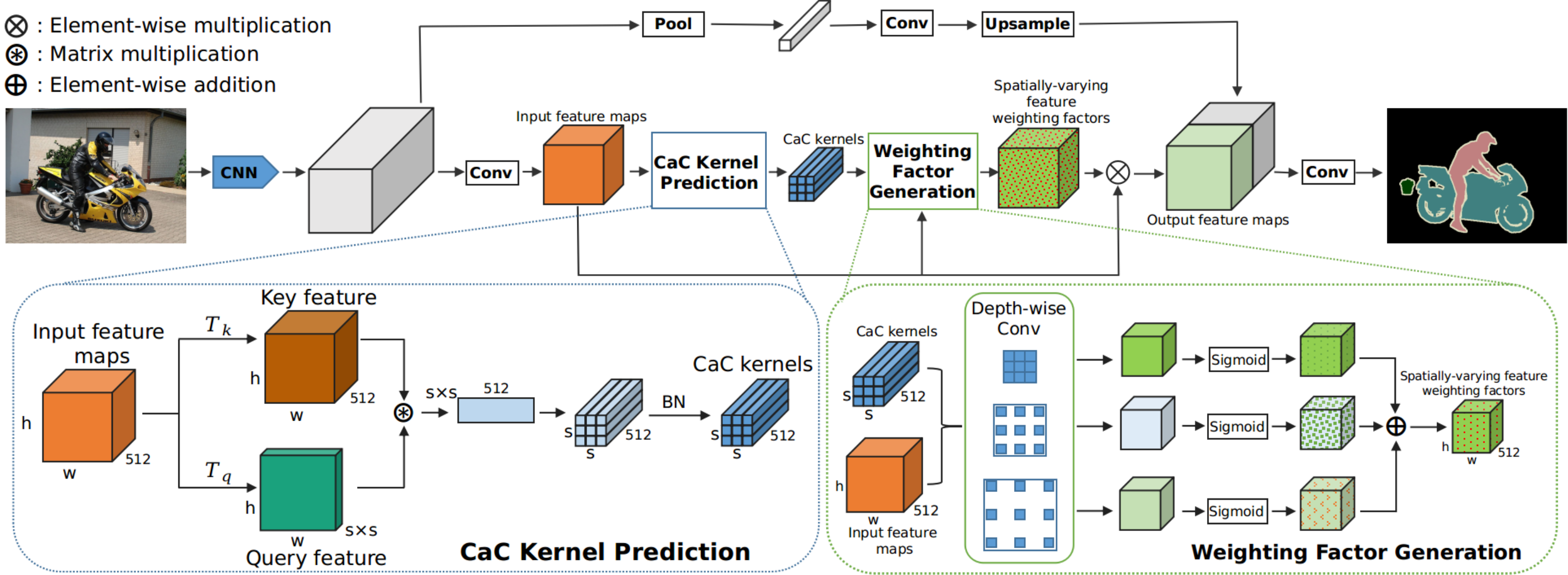}
        \end{center}
        \caption{Pipeline of Context-adaptive Convolution Network
    (CaC-Net). CaC-Net consists of a backbone convolution neural network (CNN), a
    Context-adaptive Convolution (CaC) kernel prediction module and a spatially-varying weight
generation module. The key component, CaC kernel prediction, learns to
predict a spatially-sharing context-aware convolution kernels. The weight
generation module uses the predicted dynamic kernels to generate one spatially-varying
feature weighting factor for each location in the input feature maps}
        \label{fig:pipeline}
\end{figure*}

\section{Method}
\label{sec:methods}
Feature re-weighting has proven to be a powerful approach to capture long-range semantic context by learning channel-wise weighting factors from the global contextual information.
Although it has shown its effectiveness in various tasks including
classification, detection and semantic segmentation
\cite{hu2018squeeze,Zhang_2018_CVPR}, one of its key issues is that the weighting vector is shared by all spatial locations of the 2D feature map.

We argue that, for semantic segmentation, a globally-shared weighting vector
is not an optimal solution, as different spatial locations generally belong to
objects of different classes. Therefore, desirable feature weighting factors
should still be learned from the global context but be spatially-varying to
capture different locations' unique characteristics to achieve high
segmentation accuracy. Predicting convolution kernels from the global context
to generate the spatially-varying weights is a desirable solution. Compared
with traditional convolutions on the feature maps, which only have local
receptive fields and are input-invariant, predicted kernels from the global
context are well aware of the overall scene structure for better weighting the features at different locations.

%
To achieve this goal, we propose a novel Context-adaptive Convolution (CaC) Network (see Fig. \ref{fig:pipeline}), which consists of a 2D Convolutional Neural Network (CNN)-based backbone for encoding the input images into 2D feature maps and a Context-adaptive Convolution (CaC) module for learning to channel-wisely re-weight the 2D feature maps with global context. A series of context-adaptive convolution kernels are predicted from the CaC module. 
Unlike the previous dynamic filters, which require using a large number of learnable parameters to predict the kernels, the proposed CaC kernels are predicted in a parameter-efficient manner via simple matrix multiplication. Importantly, the CaC kernels are predicted from global context of the 2D feature maps. 
The convolution with the CaC kernels can well integrate the global context as
well as the multi-scale information of the 2D feature maps to predict the spatially-varying feature re-weighting factors.
The proposed framework achieves state-of-the-art segmentation accuracy on multiple public benchmarks with small computational overhead.

\subsection{Context-adaptive Convolution Kernel Prediction}

We adopt a pre-trained ResNet \cite{he2016deep} with the dilated convolutions as the
backbone of our segmentation framework. Following \cite{yu2017dilated,chen2017deeplab,Zhang_2018_CVPR}, we remove the
downsampling operations and set the dilation rates to be $2$ and $4$ at the last two ResNet-blocks
to generate an output feature map of $1/8$ spatial size of the input image. 
It encodes each input 2D image into a 2D feature map $X \in
\mathbb{R}^{h\times w\times c}$, where $h,w,c$ are the height, width, and
feature channels of the feature map. The feature map can roughly capture
semantic information of the input image. Properly re-weighting the feature map
$X$ in a sptially-varying manner according to the global context can boost the segmentation performance.


%
To conduct spatially-varying feature re-weighting via kernel prediction, there are two naive solutions: (1) 
one can directly predict $c$ sets of $s\times s \times c$ convolutional kernels (where $s \times s$ is the kernel spatial size) following dynamic filters
\cite{jia2016dynamic} and then conducting convolution of the input feature map $X$ with the kernels to generate the $c$ sets of feature weighting maps.
However, this solution requires a fully connected (FC) layer to predict the kernel weights, 
which has too many ($s^2c^3$) learnable parameters and computation cost. 
(2) One can also predict the kernels from the global average pooled features of the input 
feature map with an FC layer to capture the global context. Feature weighting maps can be 
obtained by conducting convolution on the input feature map with the predicted kernels. 
However, such a solution loses all spatial information during kernel prediction and show inferior performance in our experiments.


%
To tackle the challenges, our proposed Context-adaptive Convolution module
predicts a series of $s\times s \times c$ CaC convolution kernels from the
global context in a parameter-efficient manner. The input feature map $X \in
\mathbb{R}^{h\times w\times c}$ is first transformed into the 2D query feature 
map $Q \in \mathbb{R}^{h\times w \times s^2}$ and the key feature map 
$K \in \mathbb{R}^{h \times w \times c}$ by two independent transformations
$T_k$ and $T_q$, respectively. 
The transformations $T_k$ and $T_q$ are independently implemented by $1\times 1$ convolutions.
Generally, the key feature $K \in \mathbb{R}^{h \times w \times c}$ captures $c$ different characteristics of the input feature map $X$ via its $c$-dimensional feature maps, while the query 
feature $Q \in \mathbb{R}^{h\times w \times s^2}$ is used to capture the global spatial distributions of $K$, where $s^2$ sets of global spatial characteristics would be captured by $Q$.

To achieve the goal, the query feature and key feature are first reshaped to obtain $\bar{Q} \in \mathbb{R}^{n\times s^2}$ and $\bar{K} \in \mathbb{R}^{n \times c}$, where $n = h \times w$. For the $i$th column $\bar{Q}(:,i) \in \mathbb{R}^{n}$ of query feature, it can be used to capture the overall spatial distribution of each feature channel $j$ of the key feature $\bar{K}(:,j) \in \mathbb{R}^n$ over all $n = h \times w$ spatial locations via the dot product $\langle Q(:,i), K(:,j) \rangle$. The result would be a scalar to measure the similarity between the spatial distributions $Q(:,i)$ and $K(:,j)$. If we repeat the procedure for all $c$ feature channels of $K$, we obtain a $c$-dimensional vector to characterize the spatial distributions of each feature channel of $\bar{K}$ with a query vector $\bar{Q}(:,i)$. Since we have $s^2$ query vectors in total, we can capture $s^2$ characteristics of the overall spatial distributions of the $c$-dimensional key feature map $K$ as
\vspace{-8pt}
\begin{align}
        \bar{D} = \bar{Q}^\top \bar{K},
        \label{eq:cac}
\end{align}
where $\bar{D} \in \mathbb{R}^{s^2 \times c}$. 

We then reshape $\bar{D}$ into the size of $s \times s \times c$ and use a
batch normalization to modulate it to obtain the predicted $D \in \mathbb{R}^{s \times s \times c}$ as our CaC kernels, which are used to convolved with the input feature map $X$ to generate the spatially-varying feature re-weighting factors for all $h\times w$ spatial locations. 

Note that the CaC kernels are predicted and would therefore be adaptive to
different inputs to capture their different global context. There are two
distinct advantages of the predicted convolution kernel $D$. 1) The CaC
kernels are able to capture the global context of the input feature map $X$,
since the matrix multiplication in Eq. \eqref{eq:cac} considers all spatial
locations for kernel prediction. 2) The CaC kernels are generated in a
parameter-efficient manner, where only $c^2 + s^2c$ learnable parameters are needed
with an inference time complexity of $O(c^2 + s^2c + sc)$ for kernel prediction, which
is significantly lower than $O(s^2c^3)$ parameters and complexity of dynamic filters.


\subsection{Spatially-varying Weight Generation}
We use the predicted CaC kernels to produce the spatially-varying weighting map for weighting each pixel of the input feature maps. 
The predicted kernels $D\in \mathbb{R}^{s\times s\times c}$ are used in a depth-wise convolution. Therefore, each channel of $D$ is responsible for modulating one channel of the input feature maps independently.
To further encourage the predicted kernel $D$ being scale invariant as well as capturing multi-scale context of the input feature map. We denote the original predicted kernels $D$ with dilation $1$ as $D_1$, and create another two CaC kernels with shared parameters of $D$ but with different dilation rates $3$ and $5$, which are denoted as $D_2$ and $D_3$.

As shown in Figure \ref{fig:pipeline}, for each set of predicted CaC kernels
of $D_1, D_2, D_3$, they are used to separately perform depth-wise convolution
on the input feature map $X$ followed by the sigmoid function. Each of them
would generate an independent spatially-varying weighting map $W_1, W_2, W_3
\in \mathbb{R}^{h \times w \times c}$, which are averaged to generate the
overall spatially-varying weighting map $W$,
\begin{align}
W = \frac{1}{3} (W_1 \oplus W_2 \oplus W_3),
\end{align}
where the symbol $\oplus$ denotes the element-wise addition.

Given $W$, we re-weight the input feature maps as $X^\star = X\odot W$, where $\odot$ represents the element-wise multiplication.
In this manner, we develop a computationally efficient way to predict the spatially-varying feature weighting factors for each spatial location of the input feature map $X$. For each predicted feature weighting vector at the spatial location $W(i,j) \in \mathbb{R}^c$, it re-weights the features $X(i,j)$ according to the global contextual information. Intuitively, the larger values in the $W(i,j)$ vector highlight more target class-related features while the smaller values suppress non-target-class-related features according to the global context.

%

\subsection{Global Pooling and Multi-head Ensembles}

Following state-of-the-art segmentation frameworks
\cite{Zhang_2019_CVPR,zhao2017pyramid}, we also integrate a global pooling
branch in our framework, which globally averages the feature vectors of all
spatial locations and replicates the vector to all spatial locations. This
replicated global-pooling feature maps are channel-wise concatenated with the
re-weighted feature maps $X^\star$ to obtain the final feature maps for segmentation.

%
%
The multi-head ensembling strategy is implemented by performing multiple proposed CaC modules in parallel and producing multiple output feature maps.  
Since the different head modules operate on the different subspace of the input feature maps,
some previous works \cite{Zhang_2019_CVPR,he2019dynamic,vaswani2017attention} show that the multi-head ensembling strategy is able to further increase the performance.
In our work, we apply this strategy to improve the capability of our CaC-Net by concatenating two output feature maps from our CaC modules.

\section{Experiments}
In this section, we first introduce the implementation details, training strategies and evaluation metrics of the 
experiments.
Then, to evaluate the proposed CaC-Net, we conduct comprehensive experiments
on three public datasets, Pascal Context \cite{mottaghi2014role}, Pascal VOC 2012 \cite{everingham2010pascal}
and ADE20K \cite{zhou2017scene}. The ablation study of our CaC-Net is carried out on the 
Pascal Context dataset. 
Finally, we report the overall results on PASCAL Context, PASCAL VOC 2012 and ADE20K.

\subsection{Implementation Details}
\noindent
\textbf{Network Structure.}\
We adopt ResNet \cite{he2016deep} as our backbone. The stride of the last two stages of
the backbone networks is removed and these dilation rates are set as $2$ and $4$
respectively. Thus the size of the feature maps  is $8\times$ smaller than
that of the input image. To predict a semantic label for each pixel, the output of our CaC-Net 
is upscaled to the size of input image by bilinear interpolation.
The ImageNet \cite{russakovsky2015imagenet} pre-trained weights are adopted to initialize the backbone networks.
%
%

%
\noindent
\textbf{Training Setting.}\
A poly learning rate policy \cite{chen2017deeplab},
 $lr = initial\_lr \times (1 - \frac{iter}{total\_iter})^{power}$ 
is used. We set the initial learning rate as $0.001$
for PASCAL Context \cite{mottaghi2014role}, $0.002$ for PASCAL VOC 2012
\cite{everingham2010pascal} and $0.004$ for
ADE20K \cite{zhou2017scene}. The power of poly learning rate policy is set as $0.9$. The optimizer
is stochastic gradient descent (SGD) \cite{bottou2010large} with momentum $0.9$ and weight
decay $0.0001$.
We train our CaC-Net for $120$ epochs for PASCAL Context dataset, $80$ epochs for PASCAL 2012 dataset and $180$ epochs for ADE20K dataset. 
We set the crop size to $512\times 512$ on PASCAL Context and PASCAL 2012. Due to the average
image size is larger than the other two datasets, we use $576 \times 576$ as the crop size on ADE20K. For data augmentation, we only randomly flip the input image and scale it randomly in the range from $0.5$ to $2.0$. 
As the prior work \cite{Zhang_2018_CVPR,Zhang_2019_CVPR}, we adopt an auxiliary segmentation loss, which is added after Res-4. 
We adopt Sync-BN \cite{Zhang_2018_CVPR} for normalization and set the batch size as 16 for all experiments.

\noindent
\textbf{Evaluation Metrics.}\
We choose the standard evaluation metrics of pixel accuracy (pixAcc) and mean Intersection of Union (mIoU) as the evaluation metrics in this experiments. Following the best practice \cite{Zhang_2018_CVPR,he2019adaptive,fu2019dual}, 
we apply the strategy of averaging the network predictions in multiple scales for evaluation. For multi-scale evaluation, we first resize the input image to multiple scales and horizontally flip them. Then the
predictions are averaged as final predictions.

\subsection{Results on PASCAL Context}
PASCAL Context dataset \cite{mottaghi2014role} is a challenging scene understanding dataset, which provides the semantic labels for the images. 
There are $4,998$ images for training and $5,105$ images for validation on PASCAL Context dataset. 
Following previous work \cite{Zhang_2018_CVPR,Zhang_2019_CVPR}, the $59$ most frequent categories are used for training our CaC-Net and all the other classes are considered as the background class. 

%

\noindent
\textbf{Ablation study on CaC-Net.}\
We conduct experiments with different settings to
evaluate the performance of our proposed CaC-Net using a ResNet-50 \cite{he2016deep}
backbone on the PASCAL Context dataset. The baseline is a ResNet-50 based
FCN \cite{long2015fully} by removing the proposed CaC modules and the
global pooling branch in our CaC-Net. 
We choose the pixAcc and mIoU for $60$ classes on the PASCAL Context dataset as
our evaluation metrics for the ablation study.

\begin{table}[t]
\begin{minipage}[t]{0.49\textwidth}
\caption{Ablation study of CaC-Net on PASCAL Context dataset. CaC represents our proposed
        CaC module. GP indicates the global pooling}
\label{table:ablation_components}
\resizebox{\textwidth}{!}{
\begin{tabular}{lllll|l}
\hline
\textbf{Method} & \textbf{Backbone} & \textbf{CaC} & \textbf{GP} &
        \textbf{pixAcc\%} & \textbf{mIoU\%} \\ \hline\hline
        FCN    &ResNet50 &  &  & 76.0 & 45.8 \\
        CaC-Net &ResNet50 & \textbf{\checkmark} &    & 79.8 & 52.0 \\
        CaC-Net &ResNet50 & \textbf{\checkmark} & \textbf{\checkmark} & 80.2 & 52.5 \\
        CaC-Net &ResNet101 & \textbf{\checkmark} & \textbf{\checkmark}& 81.5 & 55.4 \\ \hline
\end{tabular}}
\end{minipage}\hfill
\begin{minipage}[t]{0.49\textwidth}
\caption{Ablation study of different setting in training and evaluation
strategies. DS:
Deep supervision loss strategy \cite{zhao2017pyramid}. Flip: horizontally
flipping the input image for evaluation. MS: Multi-scale evaluation}
\label{table:ablation_eval}
\resizebox{\textwidth}{!}{
\begin{tabular}{ccccc|c}
\hline
        \textbf{Backbone} & \textbf{DS} & \textbf{Flip} & \textbf{MS} &
        \textbf{pixAcc\%} & \textbf{mIoU\%} \\ \hline\hline
         ResNet50 & &  &  & 79.4 & 50.7 \\
         ResNet50 & \textbf{\checkmark} &  &   & 79.7 & 51.5 \\
         ResNet50 & \textbf{\checkmark} & \textbf{\checkmark} & & 79.9 & 51.8 \\
         ResNet50 & \textbf{\checkmark} & \textbf{\checkmark} & \textbf{\checkmark}& 80.2 & 52.5 \\
\hline
\end{tabular}}
\end{minipage}
\end{table}

We first evaluate the individual components, which are added into the baseline
FCN one at a time. The size of predicted CaC kernels is $3\times 3$ and we use two CaC modules 
in all experiments. As shown in Table \ref{table:ablation_components}, 
the baseline FCN with a ResNet50 backbone achieves $76.0\%$ pixAcc and $45.8\%$ mIoU. 
With our proposed CaC, the results of pixACC and mIoU are increased by $3.8\%$ and $6.2\%$
respectively. The global pooling branch results in a further $0.5\%$ mIoU improvement. We can 
see that our proposed CaC module can significantly improve the segmentation results.
We also conduct experiments to explore training and evaluation strategies,
other plausible ways of generating the weighting maps, the influence of the number of our
proposed CaC modules for feature concatenation as introduced in Sec. \ref{sec:methods}, the
kernel size of our predicted CaC kernels, and the dilation rates in the 
feature weighting factor generation. 

\noindent
\textbf{Ablation study on training and evaluation strategies.}\
We conduct experiments to explore the effects of training and evaluation strategies and 
the results are shown in Table \ref{table:ablation_eval}. We use a deep
supervision strategy by adding an FCN head to the output of ResNet-4 as an auxiliary loss. From the table, we observe that the deep supervision training strategy results in a $0.8\%$ mIOU improvement.
We adopt the image flipping and multi-scale evaluation strategy during inference. These two evaluation strategies can further boost the performance of segmentation.

\begin{table}[t]

\begin{minipage}[t]{0.32\textwidth}
\caption{The results of different numbers of our CaC module exploited in
    the CaC-Net}
\label{table:ablation_modules}
\resizebox{\columnwidth}{!}{
\begin{tabular}{c|cccc}
\hline
  {}   & {$H=1$} & {$H=2$} & {$H=3$} & {$H=4$} \\ \hline\hline
           pixAcc & 79.9 & 80.2 & 79.9 & 80.0  \\
           mIoU   & 51.7 & 52.5 & 52.3 & 52.1 \\
\hline
\end{tabular}}
\end{minipage}
\hfill
%
\begin{minipage}[t]{0.32\textwidth}
\caption{The results of different kernel sizes used in our proposed CaC module}
\label{table:ablation_kernelsizes}
\resizebox{0.97\columnwidth}{!}{
\begin{tabular}{c|ccc}
\hline
{} & {$3+3$} & {$3+5$} & {$3+7$} \\ \hline\hline
        pixAcc & 80.2 & 80.2 & 80.1 \\
        mIoU & 52.5 & 52.3 & 52.3 \\
\hline
\end{tabular}}
\end{minipage}
\hfill
\begin{minipage}[t]{0.32\textwidth}
%
\caption{The results of different dilation rate sets for generating
    the weighting factors in CaC-Net}
\label{table:ablation_dilation}
\resizebox{\columnwidth}{!}{
\begin{tabular}{c|cccc}
\hline
        {} & {$\{1\}$} & {$\{1, 2\}$} & {$\{1,2,3\}$} & {$\{1,2,3,4\}$} \\ \hline\hline
        pixAcc &79.9 & 80.1 & 80.2 & 79.7 \\
        mIoU & 51.9 & 52.1 & 52.5 & 51.6  \\
\hline
\end{tabular}}
\end{minipage}
%
\\
\vspace{-10pt}
%

\begin{minipage}[t]{0.5\textwidth}
\caption{Results of alternative ways to generate weighting map and dynamic kernels.}
\label{table:ablation_dynamicFilters}
\resizebox{0.9\columnwidth}{!}{%
\begin{tabular}{c|ccc|cc|c}
\hline
{}   & {Fixed} & {FC} & {GAP} &{SEHead} &{EncHead} & {Ours} \\ \hline\hline
pixAcc & 79.7 & 79.8 & 79.8 & 79.7 &79.7 & 80.0  \\
mIoU   & 51.3 & 51.5 & 51.6 & 50.9 & 51.2 & 52.5 \\
\hline
\end{tabular}}
\end{minipage}
\hfill
%
\begin{minipage}[t]{0.48\textwidth}
\caption{Parameters and FLOPS of backbone and segmentation heads from APCNet \cite{he2019adaptive}, DMNet \cite{he2019dynamic}, and our CaC-Net.}
\label{table:ablation_flops}
\resizebox{\columnwidth}{!}{
\begin{tabular}{c|cccc}
\hline
{Methods} & {Backbone} & {APCNet} & {DMNet} &  {Ours}\\ \hline\hline
Parameters & 56M & +10M & +9M & +5M \\
FLOPS & 225G & +22G & +20G & +21G \\
\hline
\end{tabular}}
\end{minipage}
\end{table}

%
%
%
\noindent
\textbf{Ablation study on the alternative ways to generate weighting maps and dynamic kernels.}\
To investigate other possible ways to generate spatially-varying weighting maps, we keep our 
overall framework unchanged but replace our CaC modules with alternative approaches to generate the feature weighting maps.
The first intuitive way is to directly predict the feature weighting maps via traditional convolutions on the input feature maps, whose parameters are still learned during  training but are fixed to different inputs during testing. If our CaC kernels are replaced by the input-invariant convolutions of the same kernel sizes and different dilation rates.
Denoted by ``Fixed'' in Table \ref{table:ablation_dynamicFilters}, the performance of the fixed-kernel channel weighting prediction declines dramatically, which demonstrates that the predicted CaC kernels can generate context-aware feature weighting maps to boost the performance of semantic segmentation.

Meanwhile, there have some other possible ways to predict the dynamic filters. 
If we directly generate the dynamic kernel weights by one FC layer, the parameter overhead of this FC layer is ${h\times w \times s^2 \times c^2}$, which are about $hw$ times more parameters than our method and occupy too much GPU memory to be implemented in practice. 
If we instead adopt a depth-wise FC layer on the input feature map to generate the dynamic kernel weights. The parameters of the depth-wise FC layer reduce to ${h\times w \times s^2 \times c}$, which are the same as our CaC module. However, the mIoU on PASCAL Context decreases to $51.7\%$ (``FC'' in Table \ref{table:ablation_dynamicFilters}) with a ResNet-50 backbone, which is lower than our CaC-Net ($52.5\%$).

The third alternative way is to use the global average pooling (GAP) features
of the input feature maps to predict the dynamic kernels. In this way, the dynamic kernels can 
also capture global context. The input feature maps is first average pooled to a $c$-d vector 
followed by an FC layer to predict the $s\times s \times c$ dynamic kernels. However, 
the parameters of this scheme would be $s^2c^2$, significantly more than our CaC module's $c^2 + s^2c$ parameters. 
In addition, since the input feature map is average pooled first, the image spatial structure is lost, which limits its capacity on predicting spatially-aware dynamic kernels. If replacing our CaC kernels with such dynamic kernels from GAP features, the mIoU on PASCAL Context decreases to $51.9\%$ (``GAP'' in Table \ref{table:ablation_dynamicFilters}) on PASCAL Context.


\noindent \textbf{Ablation study on comparison with the globally-sharing feature
re-weighting methods.}\ To compare with globally-sharing feature re-weighting methods, 
we conduct two experiments by replacing the proposed CaC modules in our CaC-Net with
SE-Head \cite{hu2018squeeze} and EncHead \cite{Zhang_2018_CVPR} for globally-sharing feature re-weighting. The results are shown
in Table \ref{table:ablation_dynamicFilters}, which demonstrate the proposed
spatially-varying feature re-weighting method shows superior performance
than the globally-sharing feature re-weighting methods for semantic
segmentation.

\noindent
\textbf{Ablation study on the number of CaC modules.}\
Table \ref{table:ablation_modules} illustrates the effects of different number of CaC modules for multi-head ensemble in our proposed
CaC-Net, which shows that two CaC modules is optimal for achieving the
best performance for semantic segmentation for our designed CaC-Net.
The results are getting worse when $H$ is larger than 2 because the larger $H$ 
introduces additional parameters and model capacity, which might 
overfit the data if H is too large.
We therefore fix the number of CaC module to be $2$.

\noindent
\textbf{Ablation study on kernel size.}\
In following experiments, the influence of the kernel
size of the predicted dynamic CaC kernels is investigated. We report results
of three sets of kernel sizes, $3+3$, $3+5$, $3+7$, of the two CaC modules
respectively in Table \ref{table:ablation_kernelsizes}. We first set the kernel size of 
one CaC module as $3 \times 3$ according to the experience that $3\times 3$ kernels is
usually sufficient to obtain good performance and is computationally efficient. By increasing 
the kernel size of another CaC module from $3\times3$, $5\times5$, $7\times 7$, we observe that the performance of our proposed CaC-Net decreases slightly.
The reason might be increasing the kernel size also requires many additional FC parameters,
which causes the models to be difficult to train. To reduce the computational
cost and achieve the optimal performance, we set the kernel size of both CaC
modules to $3\times 3$.

\begin{table}[tb]
\begin{minipage}[t]{0.46\textwidth}
\caption{Segmentation results of state-of-the-art methods on PASCAL Context dataset}
\label{table:pascal-context}
\centering
\resizebox{0.96\columnwidth}{!}{
\begin{tabular}{lll}
\hline
\textbf{Method} & \textbf{Backbone} & \textbf{mIoU\%}\\\hline\hline
FCN-8S \cite{long2015fully} &  & 37.8 \\ 
CRF-RNN \cite{CRF-RNN} &  & 39.3 \\ 
ParseNet \cite{ParseNet} &  & 40.4 \\ 
HO\_CRF \cite{HO_CRF} &  & 41.3 \\ 
Piecewise \cite{Piecewise} &  & 43.3 \\ 
DeepLab-v2 \cite{chen2017deeplab} & ResNet101-COCO & 45.7 \\
RefineNet \cite{RefineNet} & ResNet152 & 47.3 \\
MSCI \cite{MSCI} & ResNet152 & 50.3 \\
EncNet \cite{Zhang_2018_CVPR} & ResNet101 & 51.7 \\
DANet \cite{fu2019dual} & ResNet101 & 52.6 \\ 
APCNet \cite{he2019adaptive} & ResNet101 & 54.7 \\ 
CFNet \cite{Zhang_2019_CVPR} & ResNet101 & 54.0 \\ 
ACNet \cite{Fu_2019_ICCV} & ResNet101 & 54.1 \\ 
APNB \cite{zhu2019asymmetric} & ResNet101 & 52.8 \\ 
DMNet \cite{he2019dynamic} & ResNet101 & 54.4 \\
\hline
Ours & ResNet50 & 52.5 \\
Ours & ResNet101 & \textbf{55.4} \\ \hline
\end{tabular}}
\end{minipage}
\hfill
%
\begin{minipage}[t]{0.46\textwidth}
\caption{Segmentation results of state-of-the-art methods on ADE20K validation set}
\label{table:ade20k}
\centering
\resizebox{0.88\columnwidth}{!}{
\begin{tabular}{lll}
\hline
\textbf{Method} & \textbf{Backbone} & \textbf{mIoU\%}\\\hline\hline
FCN \cite{long2015fully} &  & 29.39 \\
SegNet \cite{SegNet} &  & 21.64 \\
RefineNet \cite{RefineNet} & ResNet152 & 40.7 \\
PSPNet \cite{zhao2017pyramid} & ResNet101 & 43.29 \\
EncNet \cite{Zhang_2018_CVPR} & ResNet101 & 44.65 \\
SAC \cite{zhang2017scale} & ResNet101 & 44.30 \\
PSANet \cite{zhao2018psanet} & ResNet101 & 43.77 \\
UperNet \cite{zhou2017scene} & ResNet101 & 42.66 \\
APCNet \cite{he2019adaptive} & ResNet101 & 45.38 \\
CFNet \cite{Zhang_2019_CVPR} & ResNet101 & 44.89 \\
CCNet \cite{Huang_2019_ICCV} & ResNet101 & 45.22 \\ 
APNB \cite{zhu2019asymmetric} & ResNet101 & 45.24 \\ 
ACNet \cite{Fu_2019_ICCV} & ResNet101 & 45.90 \\ 
DMNet \cite{he2019dynamic} & ResNet101 & 45.50 \\ 
\hline
Ours & ResNet101 & \textbf{46.12} \\ \hline
\end{tabular}}
\end{minipage}
\end{table}

\noindent
\textbf{Ablation study on dilation rates.}\
For the spatially-varying feature weighting factor generation, we exploit a
set of dilated depth-wise convolutions in each predicted CaC kernel
to increase the ability of capturing multi-scale contextual information.
The different dilation rates are explored here. Table \ref{table:ablation_dilation} 
shows the results of different dilation rates adopted in our CaC-Net. 
We found that the dilated convolution could boost the performance of our
proposed CaC-Net as the dilated convolution increase the receptive fields.
However, the performance of the dilated rate sets $\{1,2,3,4\}$ is lower than that
of $\{1,2,3\}$ because of two aspects: (a) The receptive field of the
dilation rate $3$ is enough for the size of feature maps in our proposed CaC-Net
for the PASCAL Context dataset. (b) Too many dilation rates bring too large
computational complexity. Based on this observation, the dilation set that we
choose is $\{1,2,3\}$ in all experiments.

\begin{figure}[tb]
\centering
\subfloat{\includegraphics[width=2.39cm]{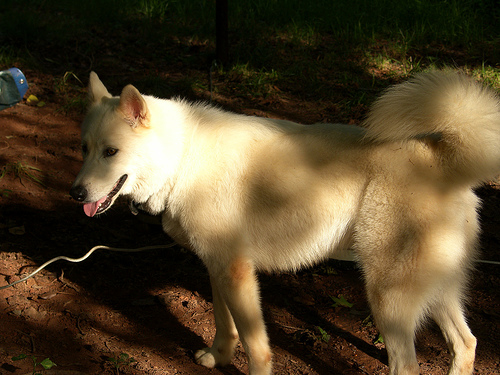}}
\hspace{0.005in}
\subfloat{\includegraphics[width=2.39cm]{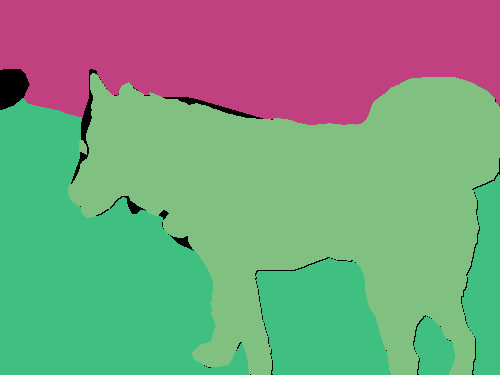}}
\hspace{0.005in}
\subfloat{\includegraphics[width=2.39cm]{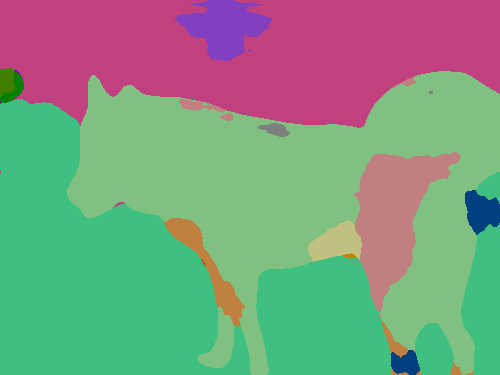}}
\hspace{0.005in}
\subfloat{\includegraphics[width=2.39cm]{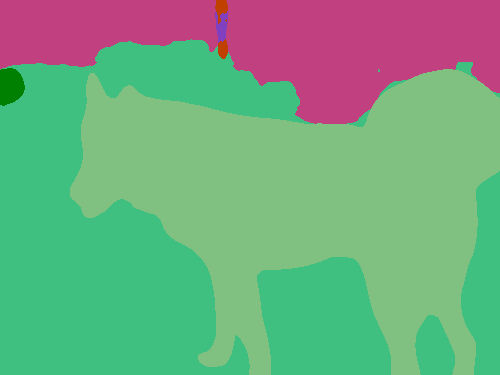}}\\
\vspace{-0.1in}
\subfloat{\includegraphics[width=2.39cm]{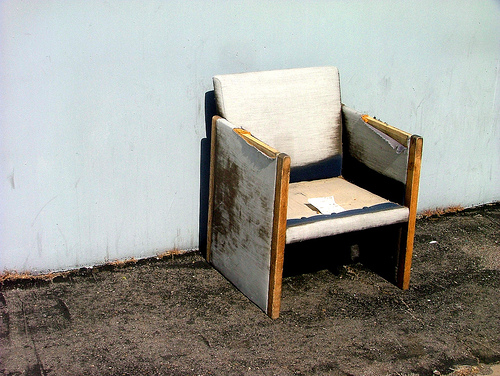}}
\hspace{0.005in}
\subfloat{\includegraphics[width=2.39cm]{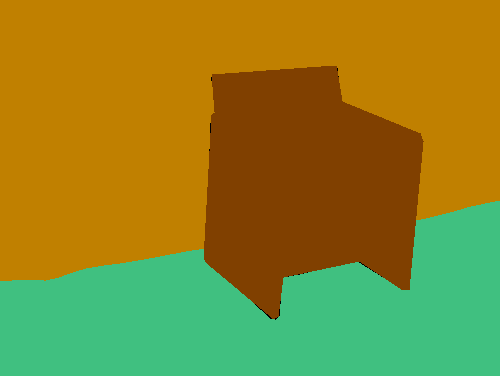}}
\hspace{0.005in}
\subfloat{\includegraphics[width=2.39cm]{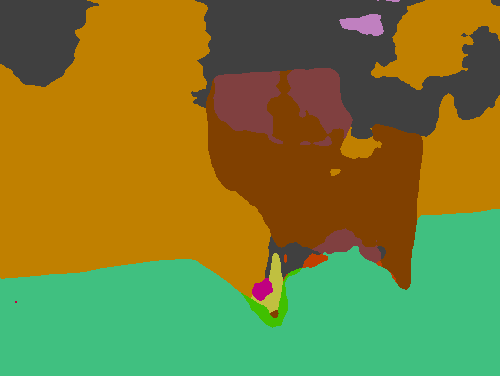}}
\hspace{0.005in}
\subfloat{\includegraphics[width=2.39cm]{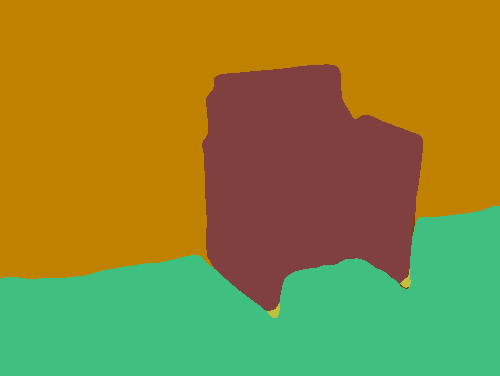}}\\
\vspace{-0.1in}
\setcounter{subfigure}{0}
\subfloat[Image]{\includegraphics[width=2.39cm]{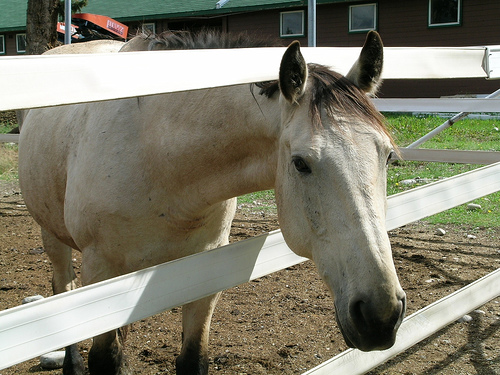}}
\hspace{0.005in}
\subfloat[GT]{\includegraphics[width=2.39cm]{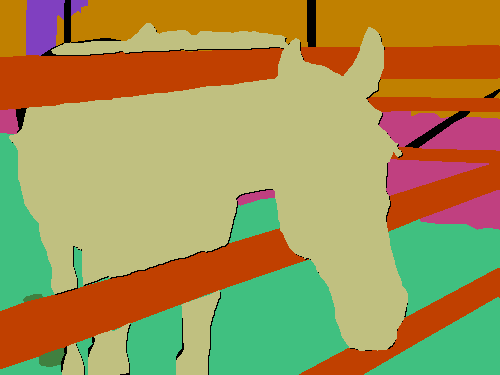}}
\hspace{0.005in}
\subfloat[Baseline]{\includegraphics[width=2.39cm]{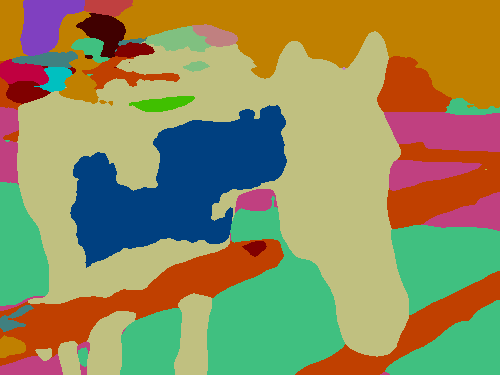}}
\hspace{0.005in}
\subfloat[CaC-Net]{\includegraphics[width=2.39cm]{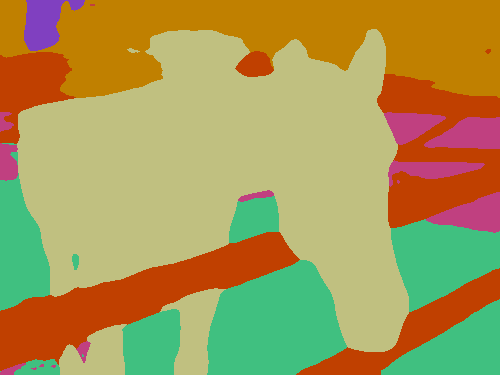}}\\
\caption{Visualization examples from the PASCAL Context dataset}
\label{fig:pcontext_vis}
\end{figure}

\noindent
\textbf{Comparisons with the state-of-the-art methods.}\
The results are shown in Table \ref{table:pascal-context}.
APCNet \cite{he2019adaptive} and CFNet \cite{Zhang_2019_CVPR} both exploit a dynamic or adaptive manner to
globally aggregate features, which has expensive computational
and memory cost.
Our proposed CaC-Net with ResNet101 backbone
surpasses other approaches by significant margins with a light-weight head.
Meanwhile, the performance of CaC-Net with ResNet50 backbone even achieves
better performance than most previous methods with a deeper backbone.
Especially, by learning to predict the spatially-varying feature weighting
factors for each location, the performance of our proposed CaC-Net
significantly outperforms EncNet \cite{Zhang_2018_CVPR}, which predicts a
globally-sharing weight factor for re-weighting feature maps.
In Figure \ref{fig:pcontext_vis}, we illustrate the visual improvements of our approach over the baseline FCN network on the PASCAL Context. It can be seen that our proposed method boosts the performance of semantic segmentation in
some challenging scenarios.

Meanwhile, with the dilated ResNet-101 backbone and input image size $512\times 512$, the
computational and parameter overheads of the backbone, the heads of APCNet,
DMNet and our CaC-Net are provided in Table \ref{table:ablation_flops}. 
Compared with the backbone, the proposed 
CaC-Net brings less than 10\% extra cost. 
Compared with the two other segmentation heads with two times more parameters, our CaC-Net has a $>1\%$ mIoU improvement on the PASCAL Context dataset.

\begin{table*}[!t]
\small
\centering
\caption{Results of each category on PASCAL VOC 2012 test set. Our
        CaC-Net achieves 85.1\% without MS COCO dataset pre-training.
    }
\label{table:pascal voc 2012}
\resizebox{\textwidth}{!}{%
\begin{tabular}{l|cccccccccccccccccccc|c}
\hline
\textbf{Method}    & \textbf{aero} & \textbf{bike} & \textbf{bird} & \textbf{boat} & \textbf{bottle} & \textbf{bus}  & \textbf{car}  & \textbf{cat}  & \textbf{chair} & \textbf{cow}  & \textbf{table} & \textbf{dog}  & \textbf{horse} & \textbf{mbike} & \textbf{person} & \textbf{plant} & \textbf{sheep} & \textbf{sofa} & \textbf{train} & \textbf{tv}   & \textbf{mIoU\%} \\ \hline\hline
\textbf{FCN} \cite{long2015fully}       & 76.8          & 34.2          & 68.9          & 49.4          & 60.3            & 75.3          & 74.7          & 77.6          & 21.4           & 62.5          & 46.8           & 71.8          & 63.9           & 76.5           & 73.9            & 45.2           & 72.4           & 37.4          & 70.9           & 55.1          & 62.2            \\
\textbf{DeepLabv2} \cite{chen2017deeplab} & 84.4          & 54.5          & 81.5
& 63.6          & 65.9            & 85.1          & 79.1          & 83.4
& 30.7           & 74.1          & 59.8           & 79.0            & 76.1           & 83.2           & 80.8            & 59.7           & 82.2           & 50.4          & 73.1           & 63.7          & 71.6            \\
\textbf{CRF-RNN} \cite{CRF-RNN}   & 87.5          & 39.0          & 79.7          & 64.2          & 68.3            & 87.6          & 80.8          & 84.4          & 30.4           & 78.2          & 60.4           & 80.5          & 77.8           & 83.1           & 80.6            & 59.5           & 82.8           & 47.8          & 78.3           & 67.1          & 72.0            \\
\textbf{DeconvNet} \cite{DeconvNet} & 89.9          & 39.3          & 79.7          & 63.9          & 68.2            & 87.4          & 81.2          & 86.1          & 28.5           & 77.0          & 62.0           & 79.0          & 80.3           & 83.6           & 80.2            & 58.8           & 83.4           & 54.3          & 80.7           & 65.0            & 72.5            \\
\textbf{DPN} \cite{DPN}       & 87.7          & 59.4          & 78.4          & 64.9          & 70.3            & 89.3          & 83.5          & 86.1          & 31.7           & 79.9          & 62.6           & 81.9          & 80.0           & 83.5           & 82.3            & 60.5           & 83.2           & 53.4          & 77.9           & 65.0            & 74.1            \\
\textbf{Piecewise} \cite{Piecewise} & 90.6          & 37.6          & 80.0          & 67.8          & 74.4            & 92            & 85.2          & 86.2          & 39.1           & 81.2          & 58.9           & 83.8          & 83.9           & 84.3           & 84.8            & 62.1           & 83.2           & 58.2          & 80.8           & 72.3          & 75.3            \\
\textbf{ResNet38} \cite{ResNet38}  & 94.4          & 72.9          & 94.9          & 68.8          & 78.4            & 90.6          & 90.0          & 92.1          & 40.1           & 90.4          & 71.7           & 89.9          & 93.7           & 91.0           & 89.1            & 71.3           & 90.7           & 61.3          & 87.7           & 78.1          & 82.5            \\
\textbf{PSPNet} \cite{zhao2017pyramid}    & 91.8          & 71.9          & 94.7          & 71.2          & 75.8            & 95.2          & 89.9          & 95.9          & 39.3           & 90.7          & 71.7           & 90.5          & 94.5           & 88.8           & 89.6            & 72.8           & 89.6           & \bgGray \textbf{64.0}          & 85.1           & 76.3          & 82.6            \\
\textbf{EncNet} \cite{Zhang_2018_CVPR}    & 94.1          & 69.2          & \bgGray\textbf{96.3} & \bgGray 76.7          & \bgGray \textbf{86.2}   & 96.3          & 90.7          & 94.2          & 38.8           & 90.7          & 73.3           & 90.0          & 92.5           & 88.8           & 87.9            & 68.7           & 92.6           & 59.0          & 86.4           & 73.4          & 82.9            
            \\
            \textbf{APCNet} \cite{he2019adaptive}      & 95.8 & 75.8 & 84.5  & 76.0 & 80.6 &
            \bgGray 96.9 & 90.0 & 96.0 & \bgGray\textbf{42.0} & \bgGray 93.7 & 75.4 & 91.6 & 95.0 & 90.5 &
            \bgGray 89.3 & \bgGray 75.8 & 92.8 & 61.9 & \bgGray 88.9 & \bgGray 79.6 & 84.2
            \\
            \textbf{CFNet} \cite{Zhang_2019_CVPR}      & 95.7 & 71.9 & 95.0  & 76.3 & \bgGray 82.8 &
            94.8 & 90.0 & 95.9 & 37.1 & 92.6 & 73.0 & \bgGray 93.4 & 94.6 & 89.6 &
            88.4 & 74.9 & \bgGray \textbf{95.2} & \bgGray 63.2 & \bgGray \textbf{89.7} & 78.2 & 84.2
            \\
            \textbf{DMNet} \cite{he2019dynamic}        & \bgGray 96.1 & \bgGray 77.3 & 94.1 & 72.8 & 78.1 & 
            \bgGray\textbf{97.1} & \bgGray \textbf{92.7} & \bgGray 96.4 & 39.8 & 91.4 & \bgGray 75.5 & 92.7 & \bgGray \textbf{95.8} &
            \bgGray {91.0} & \bgGray \textbf{90.3} & \bgGray \textbf{76.6} & \bgGray 94.1 & 62.1 & 85.5 & 77.6 & \bgGray 84.4
            \\ \hline
            \textbf{Ours}      & \bgGray \textbf{96.3} & \bgGray {76.2} &
            \bgGray 95.3 & \bgGray \textbf{78.1} & 80.8 & 96.5 
            & \bgGray {91.8}  & \bgGray \textbf{96.9} & \bgGray 40.7 & \bgGray \textbf{96.3} 
            & \bgGray \textbf{76.4} & \bgGray \textbf{94.3} & \bgGray \textbf{95.8} 
            &\bgGray \textbf{91.3} & 89.1 & 73.1  & 93.3 & 62.2 & 86.7 &
            \bgGray \textbf{80.2} & \bgGray \textbf{85.1}
   \\ \hline
\end{tabular}%
}
\end{table*}

\subsection{Results on PASCAL VOC 2012}
The PASCAL VOC 2012 dataset \cite{everingham2010pascal} is one of the most competitive semantic
segmentation benchmarks, which contains 20 foreground object classes and 1
background class. There are $1,464$ images for training, $1,449$ images
for validation and $1,456$ images for testing in the original PASCAL VOC 2012
dataset. Following the best practice \cite{Zhang_2018_CVPR,Zhang_2019_CVPR,he2019adaptive}, the augmented training set of
PASCAL VOC2012 \cite{hariharan2015hypercolumns} is adopted as our training set, which includes $10,582$
training images.
The training strategy is the same as \cite{Zhang_2018_CVPR,Zhang_2019_CVPR,he2019adaptive}, we first train our CaC-Net on the
augmented training set, and then fine-tune on the original training and validation sets.
In Table \ref{table:pascal voc 2012}, we illustrate the results of our CaC-Net
and state-of-the-art methods on PASCAL VOC 2012 benchmark. 
We can observe that our CaC-Net yields mIOU $85.1\%$ ont the test set,
which outperforms other methods without COCO pre-training
and achieves superior performance on most categories.
\begin{figure*}[tb]
\centering
\subfloat{\includegraphics[height=2.20cm]{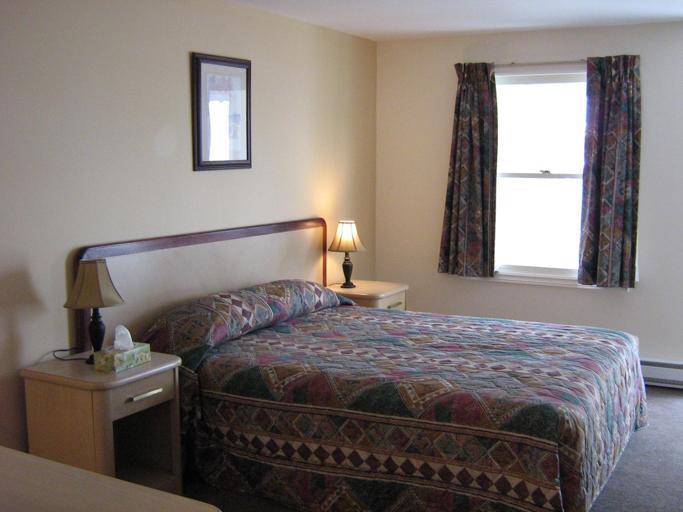}}
\hspace{0.005in}
\subfloat{\includegraphics[height=2.20cm]{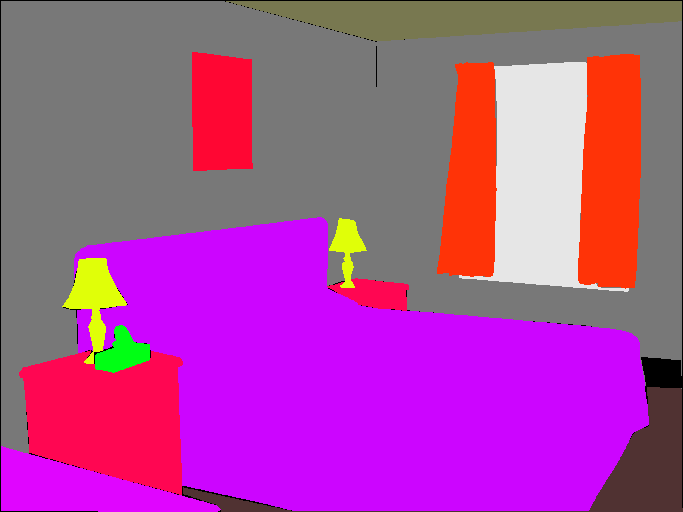}}
\hspace{0.005in}
\subfloat{\includegraphics[height=2.20cm]{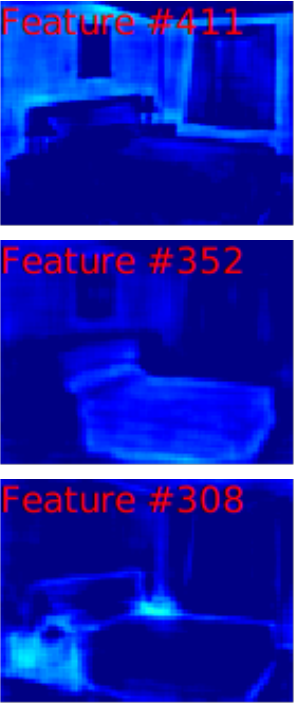}}
\hspace{0.005in}
\subfloat{\includegraphics[height=2.20cm]{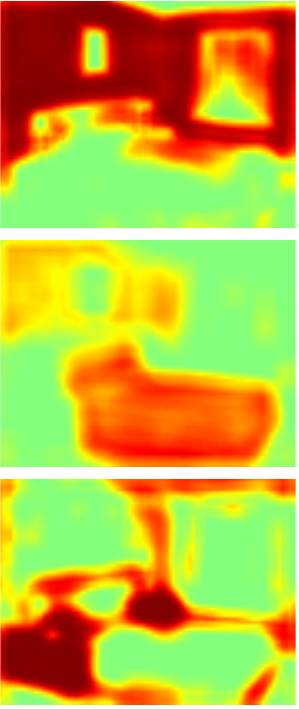}}
\hspace{0.005in}
\subfloat{\includegraphics[height=2.20cm]{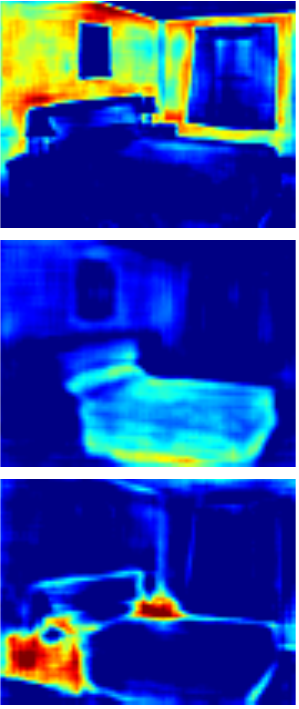}}
\hspace{0.005in}
\subfloat{\includegraphics[height=2.20cm]{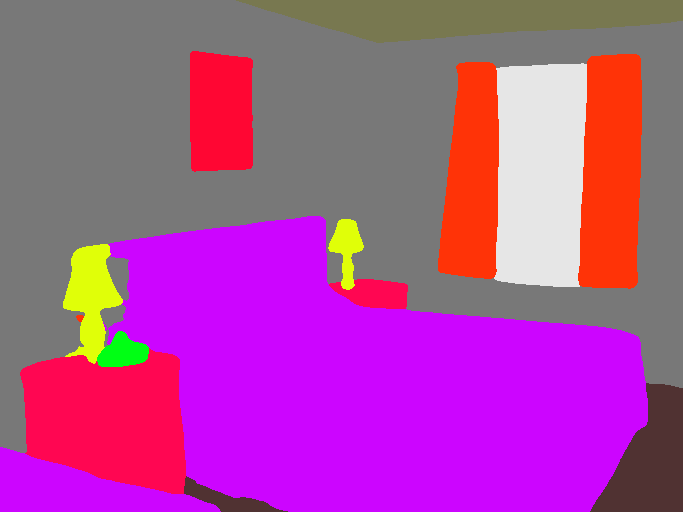}}\\
\vspace{-0.1in}
\subfloat{\includegraphics[height=2.20cm]{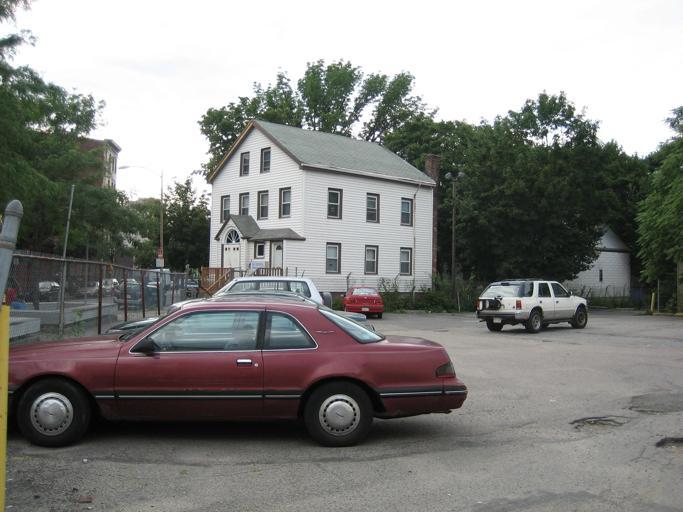}}
\hspace{0.005in}
\subfloat{\includegraphics[height=2.20cm]{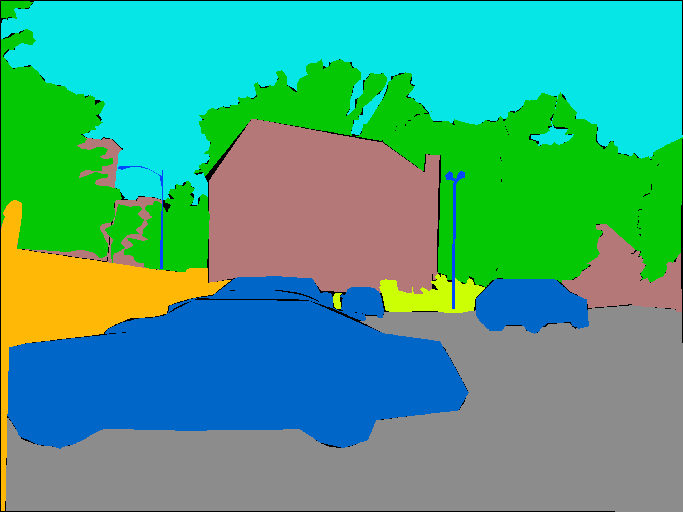}}
\hspace{0.005in}
\subfloat{\includegraphics[height=2.20cm]{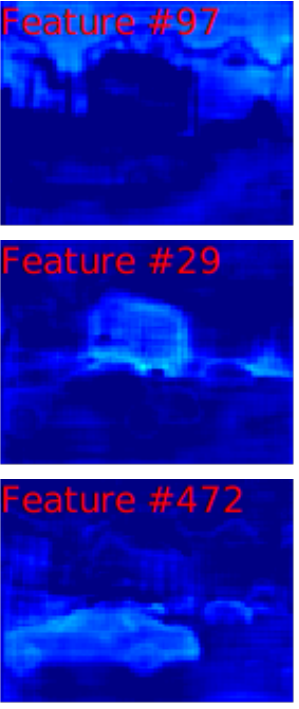}}
\hspace{0.005in}
\subfloat{\includegraphics[height=2.20cm]{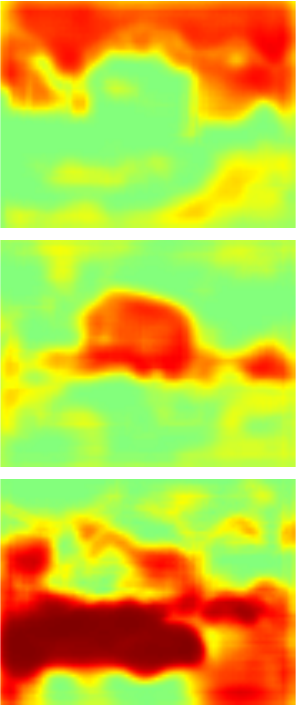}}
\hspace{0.005in}
\subfloat{\includegraphics[height=2.20cm]{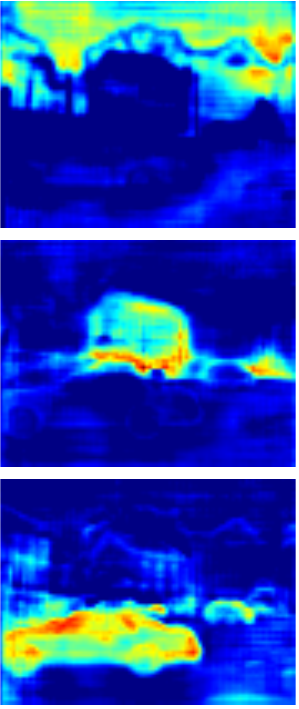}}
\hspace{0.005in}
\subfloat{\includegraphics[height=2.20cm]{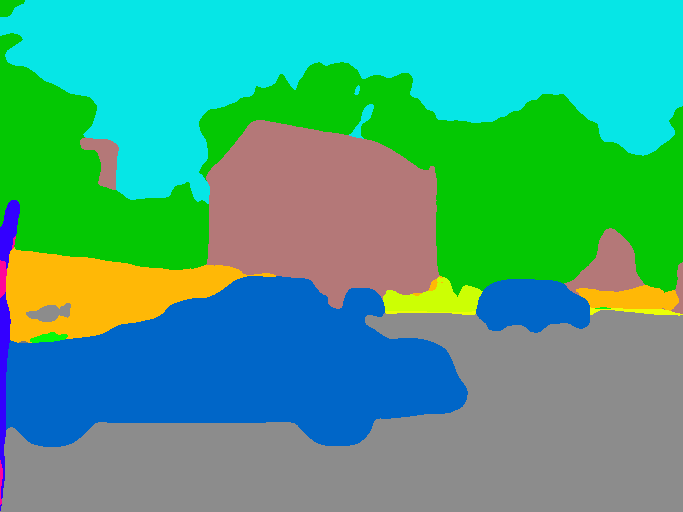}}\\
\vspace{-0.1in}
\setcounter{subfigure}{0}
\subfloat[Image]{\includegraphics[height=2.20cm]{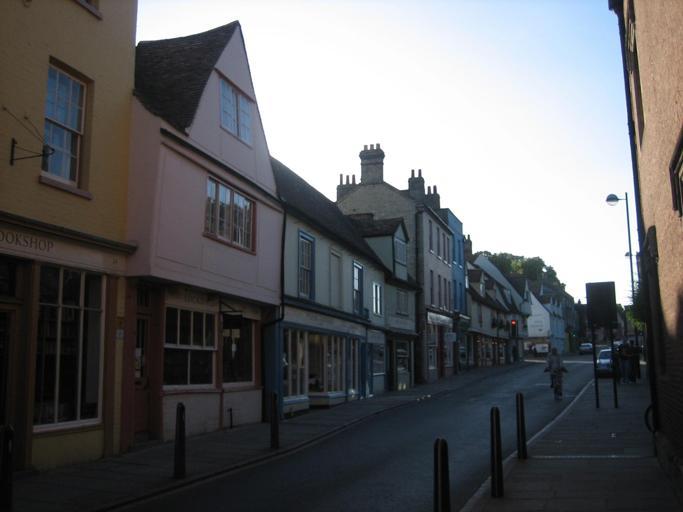}}
\hspace{0.005in}
\subfloat[Ground Truth]{\includegraphics[height=2.20cm]{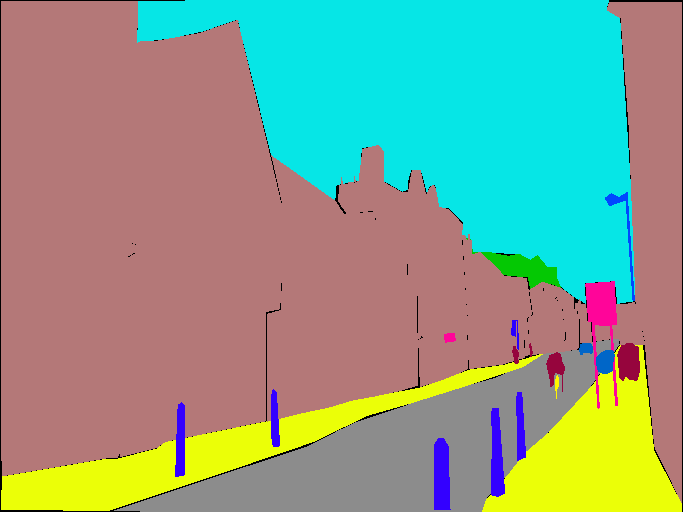}}
\hspace{0.005in}
\subfloat[\mbox{\small $X$}]{\includegraphics[height=2.20cm]{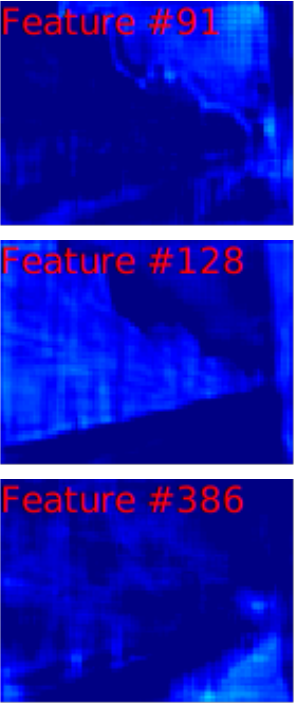}}
\hspace{0.005in}
\subfloat[\mbox{\small $W$}]{\includegraphics[height=2.20cm]{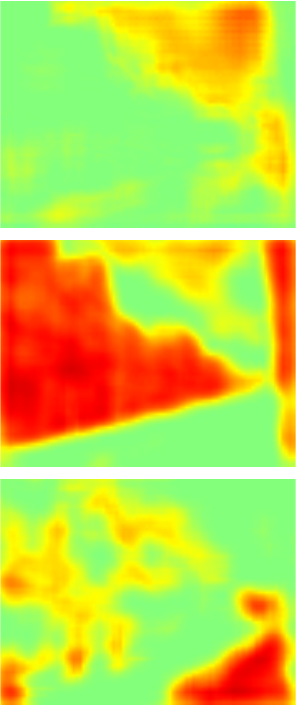}}
\hspace{0.005in}
\subfloat[\mbox{\small $X^\star$}]{\includegraphics[height=2.20cm]{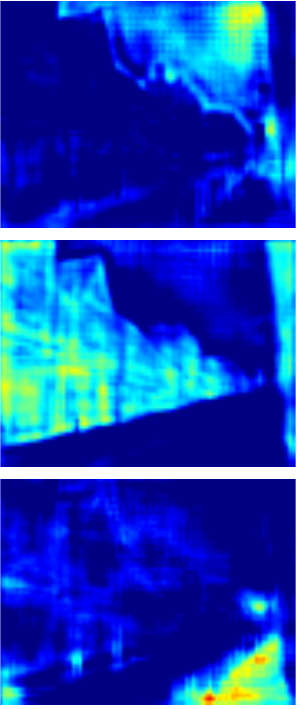}}
\hspace{0.005in}
\subfloat[CaC-Net]{\includegraphics[height=2.20cm]{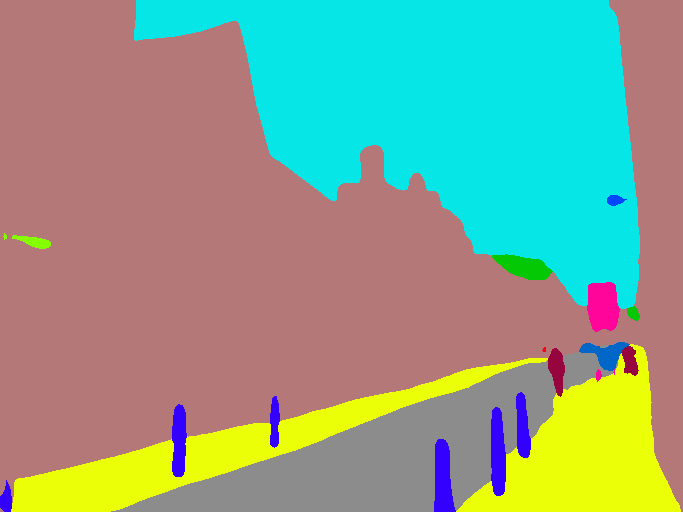}}\\
\caption{Visualization results on ADE20K dataset. (a) The input images. (b)
    Ground truth label maps. (c)-(e) Some representative
    channels of the input feature maps $X$, the output feature maps $X^\star$ and the
produced spatially-varying weighting factors $W$. In our CaC-Net, $X^\star=X\odot W $. 
(f) Results of our proposed CaC-Net}
\label{fig:ade20k_vis}
\end{figure*}

\subsection{Results on {ADE20K}}
ADE20K dataset \cite{zhou2017scene} is a large-scale scene parsing dataset which provides 
150 class categories, and consists of 20K training images, 2K validation and
3K test samples. In this subsection, we carry out experiments on ADE20K to
evaluate the effectiveness of our proposed CaC-Net.
We train our CaC-Net with a ResNet101 backbone on the training set and evaluate
the models on the validation set. Results of state-of-the-art methods on ADE20K are shown in Table \ref{table:ade20k}. Our CaC-Net with a ResNet101 backbone achieves $46.12\%$ mIOU and outperforms
all previous methods in mIoU with the same backbone.
In Figure \ref{fig:ade20k_vis}, we show some results of our
predicted results and randomly select some representative channels of their feature maps in CaC-Net. 
In the visualized feature maps, the darker colors represent the
larger values. We can observe that the spatially-varying weights produced by our CaC-Net can efficiently highlight the class-dependent regions.

\section{Conclusion}
In this paper, we propose a novel network, CaC-Net, for semantic segmentation.
The key innovation lies in the prediction of context-adaptive
convolutional kernels to integrate both global context of the input semantic feature maps. Convolution with the predicted kernels leads to feature re-weighting maps that can effectively re-weight the feature maps in a spatially-varying manner. 
%
Extensive experiments demonstrate the outstanding performance of our proposed CaC-Net, surpassing
state-of-the-art segmentation methods on multiple datasets.

%
%
{\small
\bibliographystyle{splncs04}
\bibliography{eccv2020submission_cacnet}

\begin{thebibliography}{10}
\providecommand{\url}[1]{\texttt{#1}}
\providecommand{\urlprefix}{URL }
\providecommand{\doi}[1]{https://doi.org/#1}

\bibitem{HO_CRF}
Arnab, A., Jayasumana, S., Zheng, S., Torr, P.H.: Higher order conditional
  random fields in deep neural networks. In: European Conference on Computer
  Vision. pp. 524--540. Springer (2016)

\bibitem{SegNet}
Badrinarayanan, V., Kendall, A., Cipolla, R.: Segnet: A deep convolutional
  encoder-decoder architecture for image segmentation. arXiv preprint
  arXiv:1511.00561  (2015)

\bibitem{bottou2010large}
Bottou, L.: Large-scale machine learning with stochastic gradient descent. In:
  Proceedings of COMPSTAT'2010, pp. 177--186. Springer (2010)

\bibitem{chen2017deeplab}
Chen, L.C., Papandreou, G., Kokkinos, I., Murphy, K., Yuille, A.L.: Deeplab:
  Semantic image segmentation with deep convolutional nets, atrous convolution,
  and fully connected crfs. IEEE transactions on pattern analysis and machine
  intelligence  \textbf{40}(4),  834--848 (2017)

\bibitem{chen2017rethinking}
Chen, L.C., Papandreou, G., Schroff, F., Adam, H.: Rethinking atrous
  convolution for semantic image segmentation. arXiv preprint arXiv:1706.05587
  (2017)

\bibitem{everingham2010pascal}
Everingham, M., Van~Gool, L., Williams, C.K., Winn, J., Zisserman, A.: The
  pascal visual object classes (voc) challenge. International journal of
  computer vision  \textbf{88}(2),  303--338 (2010)

\bibitem{fu2019dual}
Fu, J., Liu, J., Tian, H., Li, Y., Bao, Y., Fang, Z., Lu, H.: Dual attention
  network for scene segmentation. In: Proceedings of the IEEE Conference on
  Computer Vision and Pattern Recognition. pp. 3146--3154 (2019)

\bibitem{Fu_2019_ICCV}
Fu, J., Liu, J., Wang, Y., Li, Y., Bao, Y., Tang, J., Lu, H.: Adaptive context
  network for scene parsing. In: The IEEE International Conference on Computer
  Vision (ICCV) (October 2019)

\bibitem{hariharan2015hypercolumns}
Hariharan, B., Arbel{\'a}ez, P., Girshick, R., Malik, J.: Hypercolumns for
  object segmentation and fine-grained localization. In: Proceedings of the
  IEEE conference on computer vision and pattern recognition. pp. 447--456
  (2015)

\bibitem{he2019dynamic}
He, J., Deng, Z., Qiao, Y.: Dynamic multi-scale filters for semantic
  segmentation. In: Proceedings of the IEEE International Conference on
  Computer Vision. pp. 3562--3572 (2019)

\bibitem{he2019adaptive}
He, J., Deng, Z., Zhou, L., Wang, Y., Qiao, Y.: Adaptive pyramid context
  network for semantic segmentation. In: Proceedings of the IEEE Conference on
  Computer Vision and Pattern Recognition. pp. 7519--7528 (2019)

\bibitem{he2016deep}
He, K., Zhang, X., Ren, S., Sun, J.: Deep residual learning for image
  recognition. In: Proceedings of the IEEE conference on computer vision and
  pattern recognition. pp. 770--778 (2016)

\bibitem{hu2018squeeze}
Hu, J., Shen, L., Sun, G.: Squeeze-and-excitation networks. In: Proceedings of
  the IEEE conference on computer vision and pattern recognition. pp.
  7132--7141 (2018)

\bibitem{Huang_2019_ICCV}
Huang, Z., Wang, X., Huang, L., Huang, C., Wei, Y., Liu, W.: {CC}net:
  Criss-cross attention for semantic segmentation. In: The IEEE International
  Conference on Computer Vision (ICCV) (October 2019)

\bibitem{jia2016dynamic}
Jia, X., De~Brabandere, B., Tuytelaars, T., Gool, L.V.: Dynamic filter
  networks. In: Advances in Neural Information Processing Systems. pp. 667--675
  (2016)

\bibitem{Li_2019_ICCV}
Li, X., Zhong, Z., Wu, J., Yang, Y., Lin, Z., Liu, H.: Expectation-maximization
  attention networks for semantic segmentation. In: The IEEE International
  Conference on Computer Vision (ICCV) (October 2019)

\bibitem{MSCI}
Lin, D., Ji, Y., Lischinski, D., Cohen-Or, D., Huang, H.: Multi-scale context
  intertwining for semantic segmentation. In: Proceedings of the European
  Conference on Computer Vision (ECCV). pp. 603--619 (2018)

\bibitem{RefineNet}
Lin, G., Milan, A., Shen, C., Reid, I.: Refinenet: Multi-path refinement
  networks for high-resolution semantic segmentation. In: Proceedings of the
  IEEE conference on computer vision and pattern recognition. pp. 1925--1934
  (2017)

\bibitem{Piecewise}
Lin, G., Shen, C., Van Den~Hengel, A., Reid, I.: Efficient piecewise training
  of deep structured models for semantic segmentation. In: Proceedings of the
  IEEE Conference on Computer Vision and Pattern Recognition. pp. 3194--3203
  (2016)

\bibitem{liu2015parsenet}
Liu, W., Rabinovich, A., Berg, A.C.: Parsenet: Looking wider to see better.
  arXiv preprint arXiv:1506.04579  (2015)

\bibitem{ParseNet}
Liu, W., Rabinovich, A., Berg, A.C.: Parsenet: Looking wider to see better.
  arXiv preprint arXiv:1506.04579  (2015)

\bibitem{DPN}
Liu, Z., Li, X., Luo, P., Loy, C.C., Tang, X.: Semantic image segmentation via
  deep parsing network. In: Proceedings of the IEEE International Conference on
  Computer Vision. pp. 1377--1385 (2015)

\bibitem{long2015fully}
Long, J., Shelhamer, E., Darrell, T.: Fully convolutional networks for semantic
  segmentation. In: Proceedings of the IEEE conference on computer vision and
  pattern recognition. pp. 3431--3440 (2015)

\bibitem{mildenhall2018burst}
Mildenhall, B., Barron, J.T., Chen, J., Sharlet, D., Ng, R., Carroll, R.: Burst
  denoising with kernel prediction networks. In: Proceedings of the IEEE
  Conference on Computer Vision and Pattern Recognition. pp. 2502--2510 (2018)

\bibitem{mottaghi2014role}
Mottaghi, R., Chen, X., Liu, X., Cho, N.G., Lee, S.W., Fidler, S., Urtasun, R.,
  Yuille, A.: The role of context for object detection and semantic
  segmentation in the wild. In: Proceedings of the IEEE Conference on Computer
  Vision and Pattern Recognition. pp. 891--898 (2014)

\bibitem{niklaus2017video}
Niklaus, S., Mai, L., Liu, F.: Video frame interpolation via adaptive separable
  convolution. In: Proceedings of the IEEE International Conference on Computer
  Vision. pp. 261--270 (2017)

\bibitem{DeconvNet}
Noh, H., Hong, S., Han, B.: Learning deconvolution network for semantic
  segmentation. In: Proceedings of the IEEE international conference on
  computer vision. pp. 1520--1528 (2015)

\bibitem{russakovsky2015imagenet}
Russakovsky, O., Deng, J., Su, H., Krause, J., Satheesh, S., Ma, S., Huang, Z.,
  Karpathy, A., Khosla, A., Bernstein, M., et~al.: Imagenet large scale visual
  recognition challenge. International journal of computer vision
  \textbf{115}(3),  211--252 (2015)

\bibitem{su2019pixel}
Su, H., Jampani, V., Sun, D., Gallo, O., Learned-Miller, E., Kautz, J.:
  Pixel-adaptive convolutional neural networks. In: Proceedings of the IEEE
  Conference on Computer Vision and Pattern Recognition. pp. 11166--11175
  (2019)

\bibitem{takikawa2019gated}
Takikawa, T., Acuna, D., Jampani, V., Fidler, S.: Gated-scnn: Gated shape cnns
  for semantic segmentation. In: Proceedings of the IEEE International
  Conference on Computer Vision. pp. 5229--5238 (2019)

\bibitem{vaswani2017attention}
Vaswani, A., Shazeer, N., Parmar, N., Uszkoreit, J., Jones, L., Gomez, A.N.,
  Kaiser, {\L}., Polosukhin, I.: Attention is all you need. In: Advances in
  neural information processing systems. pp. 5998--6008 (2017)

\bibitem{ResNet38}
Wu, Z., Shen, C., Hengel, A.v.d.: Wider or deeper: Revisiting the resnet model
  for visual recognition. arXiv preprint arXiv:1611.10080  (2016)

\bibitem{yu2017dilated}
Yu, F., Koltun, V., Funkhouser, T.: Dilated residual networks. In: Proceedings
  of the IEEE conference on computer vision and pattern recognition. pp.
  472--480 (2017)

\bibitem{Zhang_2018_CVPR}
Zhang, H., Dana, K., Shi, J., Zhang, Z., Wang, X., Tyagi, A., Agrawal, A.:
  Context encoding for semantic segmentation. In: The IEEE Conference on
  Computer Vision and Pattern Recognition (CVPR) (June 2018)

\bibitem{Zhang_2019_CVPR}
Zhang, H., Zhang, H., Wang, C., Xie, J.: Co-occurrent features in semantic
  segmentation. In: The IEEE Conference on Computer Vision and Pattern
  Recognition (CVPR) (2019)

\bibitem{zhang2017scale}
Zhang, R., Tang, S., Zhang, Y., Li, J., Yan, S.: Scale-adaptive convolutions
  for scene parsing. In: Proceedings of the IEEE International Conference on
  Computer Vision. pp. 2031--2039 (2017)

\bibitem{zhao2017pyramid}
Zhao, H., Shi, J., Qi, X., Wang, X., Jia, J.: Pyramid scene parsing network.
  In: Proceedings of the IEEE conference on computer vision and pattern
  recognition. pp. 2881--2890 (2017)

\bibitem{zhao2018psanet}
Zhao, H., Zhang, Y., Liu, S., Shi, J., Change~Loy, C., Lin, D., Jia, J.:
  {PSAN}et: Point-wise spatial attention network for scene parsing. In:
  Proceedings of the European Conference on Computer Vision (ECCV). pp.
  267--283 (2018)

\bibitem{CRF-RNN}
Zheng, S., Jayasumana, S., Romera-Paredes, B., Vineet, V., Su, Z., Du, D.,
  Huang, C., Torr, P.H.: Conditional random fields as recurrent neural
  networks. In: Proceedings of the IEEE international conference on computer
  vision. pp. 1529--1537 (2015)

\bibitem{zhou2017scene}
Zhou, B., Zhao, H., Puig, X., Fidler, S., Barriuso, A., Torralba, A.: Scene
  parsing through ade20k dataset. In: Proceedings of the IEEE conference on
  computer vision and pattern recognition. pp. 633--641 (2017)

\bibitem{zhu2019asymmetric}
Zhu, Z., Xu, M., Bai, S., Huang, T., Bai, X.: Asymmetric non-local neural
  networks for semantic segmentation. In: Proceedings of the IEEE International
  Conference on Computer Vision. pp. 593--602 (2019)

\end{thebibliography}
}

\end{document}